\documentclass{article}
\usepackage{spconf}
\usepackage{amsmath,amssymb,amsfonts}
\usepackage{graphicx}
\usepackage{xcolor}
\usepackage{booktabs}
\usepackage{multirow}
\usepackage{diagbox}
\usepackage[colorlinks,linkcolor=blue]{hyperref}
\usepackage{soul}
\usepackage{makecell}
\usepackage{bbding}
\usepackage{pifont}
\usepackage{bigstrut}
\usepackage{color}
\usepackage{algorithm}
\usepackage{algpseudocode}

\usepackage{amsmath}
\usepackage{amssymb}
\usepackage{algorithm}
\usepackage{algpseudocode}
\usepackage{xcolor}
\usepackage{tabularx}
\newcolumntype{C}{>{\centering\arraybackslash}X}
\usepackage{tabularx}
\usepackage{array}
\usepackage{booktabs}
\usepackage{multirow}
\newcolumntype{V}{>{\centering\arraybackslash}p{0.18\columnwidth}}
\usepackage{xurl}
\title{SRPR-Net: Semantic and Relational Prompt Refinement for Automated SAM-based Instance Segmentation}

\name{%
Lufei Liu\textsuperscript{1},
Guojie Li\textsuperscript{1},
Suncheng Xiang\textsuperscript{2,*},
Fan Zhang\textsuperscript{1,*}
\thanks{\textsuperscript{*} indicates the corresponding authors. This work was  partially supported by the Ruijin-XJTLU Medical-Engineering Interdisciplinary Research under Grant No.2026JC028 and the National Natural Science Foundation of China under Grant No.62301315.}
}

\address{%
\textsuperscript{1}Ruijin-XJTLU Intelligent Medicine Institute, Ruijin Hospital,\\
Shanghai Jiao Tong University School of Medicine, Shanghai, China;\\
and Xi'an Jiaotong-Liverpool University, Suzhou, China\\
\textsuperscript{2}School of Biomedical Engineering, Shanghai Jiao Tong University, Shanghai, China
}

\begin{document}
\setlength{\parskip}{0pt}
\maketitle
\begin{abstract}
Instance segmentation is a fundamental computer vision task with diverse real-world applications. Recently, prompt-driven foundation models have shown promising generalization. However, automated prompting remains limited by insufficient semantic guidance and inter-instance modeling. To address this challenge, we propose a novel architecture, named Semantic Relational Prompt Refinement Network (SRPR-Net), for automated SAM-based instance segmentation. 
A sequential prompt refinement mechanism is introduced to enrich detector geometry with visual-language semantics and then incorporate same-image instance dependencies, enabling context-aware box adjustment before SAM segmentation. Experiments on multiple standard benchmarks demonstrate that SRPR-Net achieves consistent improvements in segmentation performance over existing state-of-the-art approaches. The code is publicly available at \url{https://github.com/JeremyXSC/SRPR-Net}.
\end{abstract}


\section{Introduction}
\label{sec:introduction}
Instance segmentation identifies individual objects and generates pixel-level masks. Recently, vision foundation models have introduced prompt-driven segmentation. Among them, Segment Anything Model (SAM) enables flexible zero-shot segmentation using point and box prompts \cite{kirillov2023segment}. Subsequent studies improve mask quality through HQ-SAM \cite{ke2023hq}, reduce computation through EfficientSAM \cite{xiong2024efficientsam}, and extend SAM to medical imaging \cite{ma2024medsam, wu2025medicalsamadapter}. However, automated deployment requires prompts without manual spatial guidance, motivating automatic prompt generation \cite{aopsam,semanticautosam}. RSPrompter \cite{chen2024rsprompter}, USIS-SAM \cite{lian2024usis}, and UN-SAM \cite{chen2025unsam} introduce learned prompts or mask hints across different domains, while Crowd-SAM improves prompting in crowded scenes \cite{cai2024crowdsam}. Object detectors further provide automatic box prompts for SAM: YOLO-SAM \cite{chen2025unsam} combines YOLO-World \cite{cheng2024yoloworld} with EfficientSAM \cite{xiong2024efficientsam}, whereas BLO-Inst learns segmentation-oriented boxes through bi-level optimization \cite{bloinst2026}. Nevertheless, two limitations remain in box prompt refinement: 1) \textbf{Semantic representation:} box coordinates and confidence lack explicit global visual context and category semantics; 2) \textbf{Relational refinement:} independent refinement overlooks inter-instance interactions and contextual dependencies \cite{relationnet}. Together, these limitations hinder effective prompting in complex scenes, motivating context-aware refinement inspired by vision-language learning \cite{coop} and dense prediction \cite{denseclip}.

Our key innovation lies in the Semantic Relational Prompt Refinement Network (SRPR-Net), which combines CLIP-based semantic enhancement \cite{radford2021learning} with Transformer-based relational refinement \cite{vaswani2017attention}. Unlike refinement based only on box geometry and confidence, SRPR-Net first incorporates frozen global visual and category-specific text features into instance tokens. Then the relational module models dependencies among these tokens to predict segmentation-oriented box adjustments for SAM. By coupling semantic cues with
inter-instance dependencies, SRPR-Net improves prompt re
finement across diverse visual domains.

In summary, the main contributions are as follows: 1) We propose SRPR-Net, a novel framework for refining detector-generated box prompts before SAM-based instance segmentation. 2) A sequential prompt refinement mechanism is introduced to integrate visual-language semantics with inter-instance dependencies for context-aware box adjustment. 3) Extensive experiments show that SRPR-Net outperforms the baseline across multiple metrics and diverse visual domains.

\begin{figure*}[t]
    \centering
    \includegraphics[width=\linewidth]{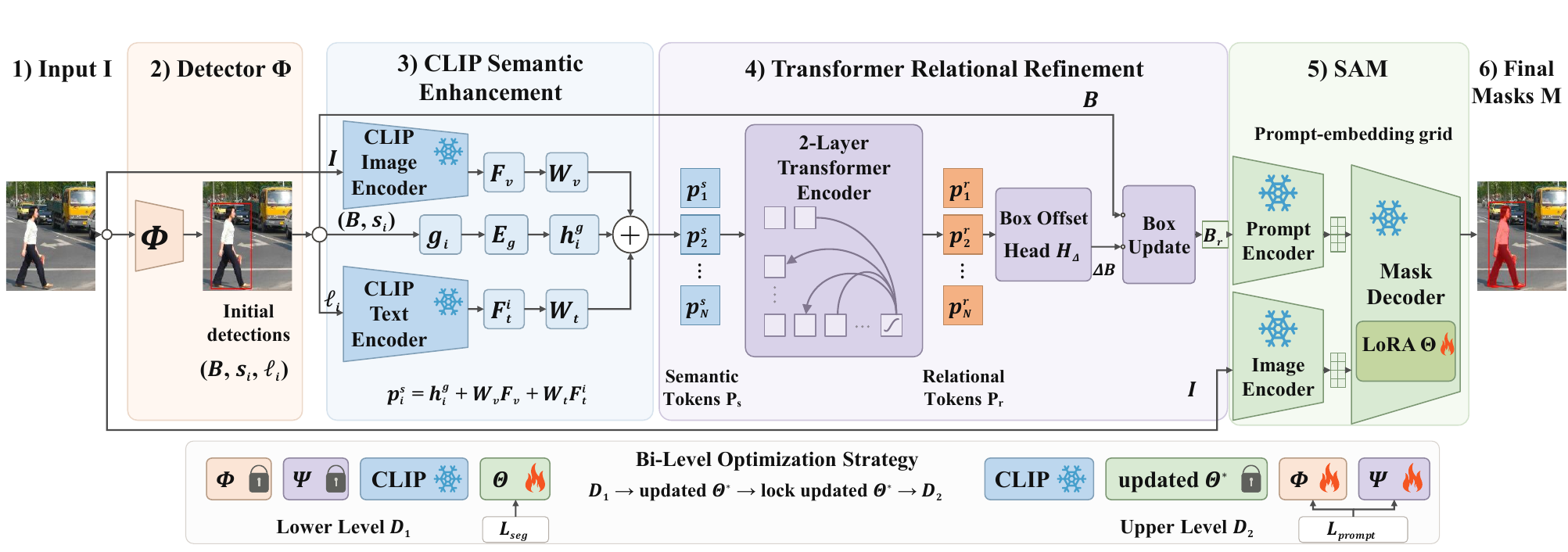}
    \caption{The overall architecture of our proposed SRPR-Net, which consists of a CLIP-based semantic prompt enhancement module and a Transformer-based relational prompt refinement module for SAM-based instance segmentation.}
    \label{fig:framework}
\end{figure*}

\section{Methodology}
\label{sec:method}
\subsection{Preliminary}
\label{subsec:preliminary}
Given an input image $I$, the detector generates a set of initial bounding box prompts $B=\{b_i\}_{i=1}^{N}$, where $N$ denotes the number of detected instances and $b_i$ corresponds to the $i$-th instance, with $i\in\{1,2,\ldots,N\}$. Using $b_i$ as a spatial prompt, SAM predicts the corresponding mask $M_i=\operatorname{SAM}(I,b_i)$ \cite{kirillov2023segment}. Since detector prompts mainly encode object location, scale, and confidence, SRPR-Net enriches them with semantic and relational information before mask generation. In this paper, $\Phi$ denotes the trainable detector parameters, $\Psi$ denotes all trainable parameters of the SRPR prompt refiner, including the geometric encoder, CLIP projection layers, Transformer encoder, and box offset head, and $\Theta$ denotes only the trainable LoRA parameters in the SAM mask decoder \cite{hu2022lora}.The CLIP encoders and all remaining SAM parameters are frozen.

\subsection{Our Proposed SRPR-Net Framework}
\label{subsec:framework}

The overall architecture of SRPR-Net is illustrated in
Fig.~\ref{fig:framework} and primarily consists of
(1) \textbf{a Semantic Prompt Enhancement Module} and
(2) \textbf{a Relational Prompt Refinement Module}. Specifically, an RGB image $I\in\mathbb{R}^{H\times W\times3}$ is processed by the detector, producing $N$ bounding boxes, confidence scores, and category labels. The boxes are represented by $B\in\mathbb{R}^{N\times4}$, with one confidence score and category label associated with each instance.

To effectively enrich the detector prompts, the frozen CLIP image encoder resizes the image to $224\times224\times3$ and extracts an L2-normalized 512-dimensional global visual feature. Each predicted category is converted into a class prompt, such as ``a photo of person,'' and encoded by the frozen CLIP text encoder into a 512-dimensional text feature \cite{radford2021learning}. Learnable projections map both CLIP features from 512 to 256 dimensions, while a geometric encoder maps the five-dimensional box and confidence description to the same dimension. Their fusion yields $N$ semantic prompt tokens $P_s\in\mathbb{R}^{N\times256}$. A two-layer Transformer encoder with eight-head self-attention then models interactions among these tokens and produces $P_r\in\mathbb{R}^{N\times256}$ \cite{vaswani2017attention}.

Finally, a lightweight box offset head maps $P_r$ to four adjustment parameters per instance, producing $\Delta B\in\mathbb{R}^{N\times4}$. These adjustments update the initial boxes to obtain $B_r\in\mathbb{R}^{N\times4}$, which is passed to SAM for final mask prediction. During training, the Lower Level adapts the mask-decoder LoRA parameters $\Theta$ using the current prompts, whereas the Upper Level updates the detector $\Phi$ and prompt refiner $\Psi$ using detector supervision and segmentation feedback on an independent data subset. This bi-level optimization strategy follows BLO-Inst \cite{bloinst2026} to reduce alignment overfitting between prompt generation and segmentation adaptation, as summarized in Algorithm~\ref{alg:optimization}.

\subsection{Semantic Relational Prompt Refinement Mechanism}
\label{subsec:refinement}

\begin{algorithm}[!t]
\caption{Optimization Process of SRPR-Net}
\label{alg:optimization}
\begin{algorithmic}[1]
\Require Detector $\Phi$, refiner $\Psi$, SAM LoRA $\Theta$, dataset $D$
\State Split $D$ into non-overlapping $D_1$ and $D_2$ (1:1)
\For{$t=0$ to $T-1$}
    \State Sample $\mathcal{B}_1\sim D_1$ and $\mathcal{B}_2\sim D_2$
    \Statex \hspace{\algorithmicindent}
    \textcolor{blue}{\textit{// Lower Level: Update SAM LoRA}}
    \State Update $\Theta$ to $\Theta^{*}$ using
    $\mathcal{L}_{\mathrm{seg}}(\Theta;\Phi,\Psi,\mathcal{B}_1)$,
    \Statex \hspace{\algorithmicindent}
    \textcolor{blue}{\textit{// Upper Level: Update Detector and Refiner}}
    \State Update $(\Phi,\Psi)$ using
    $\mathcal{L}_{\mathrm{prompt}}(\Theta^{*};\Phi,\Psi,\mathcal{B}_2)$,
\EndFor
\Ensure Optimized $\Phi^{*}$, $\Psi^{*}$, and $\Theta^{*}$
\end{algorithmic}
\end{algorithm}

On the basis of SRPR-Net, we perform semantic enhancement followed by relational refinement.

\textbf{Semantic Prompt Enhancement.}
For each detection, we construct
$g_i=[c_x,c_y,w,h,s_i]$,
where the box-center coordinates $(c_x,c_y)$ and box dimensions
$(w,h)$ are normalized by the corresponding image dimensions,
and $s_i$ is the detection confidence.
A geometric encoder comprising a linear layer, LayerNorm,
and GELU maps this descriptor to
$h_i^g=E_g(g_i)\in\mathbb{R}^{256}$.
The frozen CLIP encoders provide $\ell_2$-normalized visual
and text features $F_v,F_t^i\in\mathbb{R}^{512}$~\cite{radford2021learning}.
The global visual feature $F_v$ is shared across detections,
while $F_t^i$ encodes a class prompt derived from the predicted
category $\ell_i$.
Learnable projections $W_v$ and $W_t$ map these features
to the prompt space, yielding the semantic token
$p_i^s=h_i^g+W_vF_v+W_tF_t^i$.
Stacking these tokens forms $P_s$.

\textbf{Relational Prompt Refinement.}
A two-layer Pre-LN Transformer encoder with eight attention
heads models inter-instance relations in $P_s$
\cite{vaswani2017attention,xiong2020layernorm}.
For head $h$, queries, keys, and values are computed as
$Q_h=\hat{P}W_h^Q$, $K_h=\hat{P}W_h^K$, and
$V_h=\hat{P}W_h^V$, where $\hat{P}$ denotes the normalized
input tokens of the current layer, initialized from $P_s$,
and $W_h^Q$, $W_h^K$, and $W_h^V$ are learnable
projection matrices.
Self-attention aggregates information across instances as
\begin{equation}
Z_h=\operatorname{Softmax}
\left(\frac{Q_hK_h^\top}{\sqrt{d_k}}\right)V_h,
\label{eq:semantic_relational}
\end{equation}
where $d_k=32$.
The two encoder layers integrate self-attention,
residual connections, and feed-forward networks to produce
$P_r=\operatorname{Transformer}(P_s)\in\mathbb{R}^{N\times256}$.
A lightweight box offset head predicts
$\Delta B=H_{\Delta}(P_r)\in\mathbb{R}^{N\times4}$,
with each row specifying center and log-scale adjustments
following the standard box-regression parameterization
\cite{girshick2015fast}.

\textbf{Bi-Level Optimization.}
Following BLO-Inst~\cite{bloinst2026}, we split the training
dataset $D$ into disjoint subsets $D_1$ and $D_2$ (1:1) and
alternate optimization to mitigate alignment overfitting.
At the Lower Level, we update the SAM mask-decoder LoRA
parameters $\Theta$ by minimizing
$\mathcal{L}_{\mathrm{seg}}(\Theta;\Phi,\Psi,D_1)$,
with the detector $\Phi$ and refiner $\Psi$ fixed,
and denote the resulting parameters by $\Theta^{*}$.
At the Upper Level, we hold $\Theta^{*}$ fixed
and optimize $(\Phi,\Psi)$ by minimizing
$\mathcal{L}_{\mathrm{prompt}}(\Theta^{*};\Phi,\Psi,D_2)$,
which combines detector supervision with segmentation feedback.
The CLIP encoders and all remaining SAM parameters stay frozen
throughout training.

\section{Experiments}
\label{sec:experiments}
\subsection{Experimental Settings}
\label{sec:experimental_settings}

\begin{figure}[!t]
    \centering
    \includegraphics[width=\columnwidth]{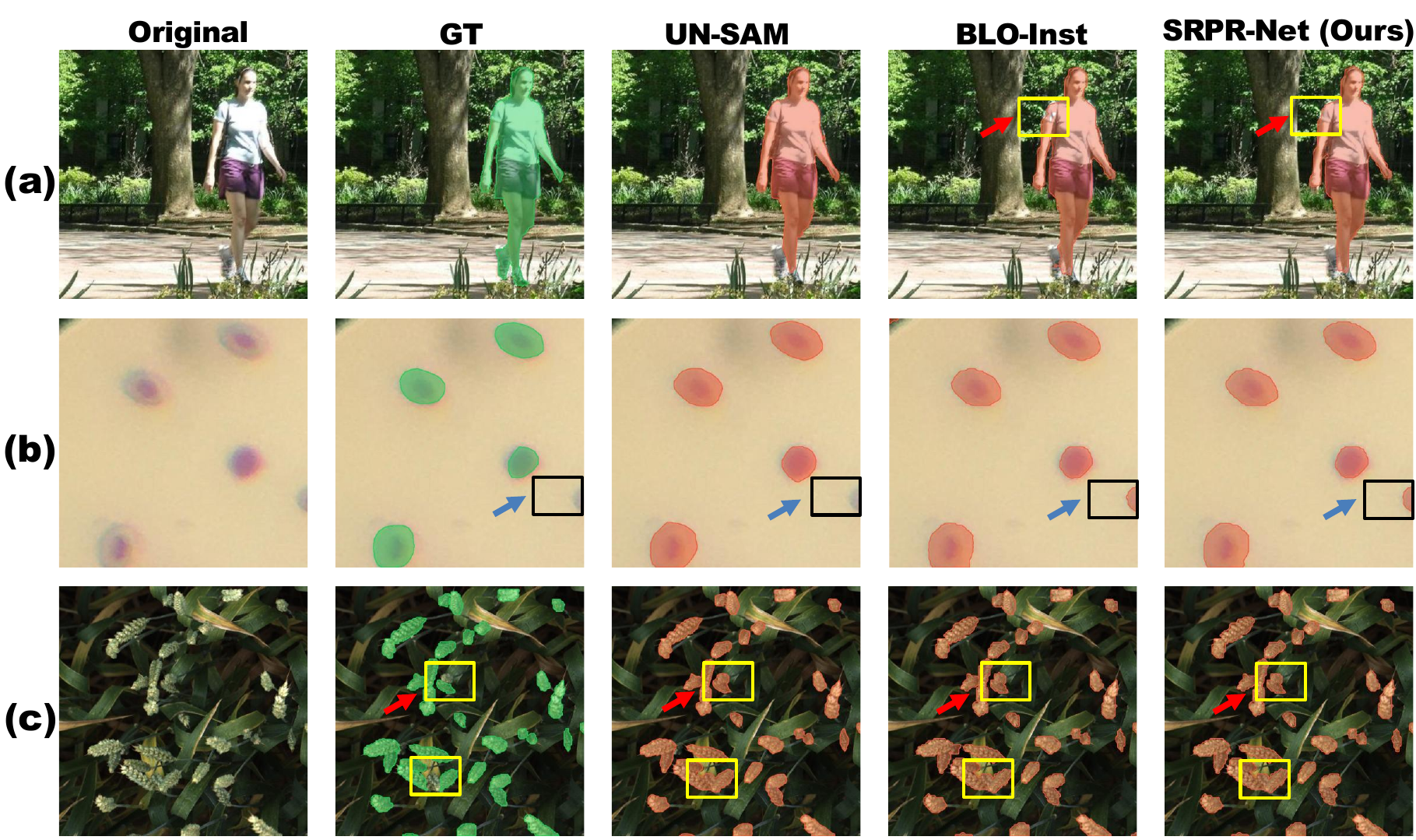}
    \caption{Qualitative comparison across three instance
    segmentation domains. Rows (a)--(c) correspond to
    PennFudanPed, RWCellIns, and WheatIns.}
    \label{fig:sota_qualitative}
\end{figure}

\textbf{Datasets.}
We evaluate SRPR-Net on three public instance segmentation
benchmarks from distinct visual domains:
PennFudanPed~\cite{wang2007object} for pedestrians,
RWCellIns~\cite{atri2023rwcellins} for microscopic blood cells,
and WheatIns~\cite{shehadeh2024wheatins} for field wheat.

\textbf{Metrics.}
Following the COCO protocol~\cite{lin2014microsoft},
we report mask mAP averaged over IoU thresholds of
$0.50{:}0.05{:}0.95$, and AP$_{50}$ and AP$_{75}$ at IoU
thresholds of 0.50 and 0.75, respectively.

\textbf{Implementation Details.}
We first pretrain the YOLO detector for 50 epochs on the training split of each target dataset. Following the bi-level optimization strategy of BLO-Inst ~\cite{bloinst2026}, we then train SRPR-Net for 20 epochs with a batch size of 1 using SGD with Nesterov momentum of 0.937.  We use SAM ViT-B as the segmentation backbone. Besides, the detector, prompt refiner, and SAM LoRA parameters are optimized with the same initial learning rate of \(10^{-3}\), which decays linearly to \(10^{-4}\) over the bi-level training stage. In addition, weight decay is set to \(5\times10^{-4}\) for the detector and prompt refiner and to zero for SAM LoRA, whose rank is set to 4. All experiments are performed using PyTorch 1.13.1 and torchvision 0.14.1 with CUDA 11.7 on an NVIDIA GeForce RTX 3090 GPU, with a fixed random seed of 2026.

\subsection{Comparison with State-of-the-Art Methods}
\label{sec:sota_comparison}

In this section, we compare our SRPR-Net with representative state-of-the-art methods in two
categories:
(1) specialist instance segmentation methods, including
Mask R-CNN~\cite{he2017mask} and SOLO~\cite{wang2021solo};
(2) SAM-based methods, including SAM+B, SAM+M,
RSPrompter~\cite{chen2024rsprompter} (anchor/query variants),
USIS~\cite{lian2024usis}, UN-SAM~\cite{chen2025unsam},
and BLO-Inst~\cite{bloinst2026}.
SAM+B and SAM+M use box prompts derived from Mask R-CNN~\cite{he2017mask}
and Mask2Former~\cite{cheng2022masked} respectively.
Our direct baseline is BLO-Inst~\cite{bloinst2026}.
As shown in Tables~\ref{tab:sota_penn}
and~\ref{tab:sota_rw_wheat}, SRPR-Net achieves the highest
mAP and AP$_{75}$ among the compared methods on all three
benchmarks.
Fig.~\ref{fig:sota_qualitative} provides qualitative
comparisons with UN-SAM and BLO-Inst across the three datasets.

\textbf{PennFudanPed.}
Compared with BLO-Inst~\cite{bloinst2026}, SRPR-Net improves
mAP by \textbf{+4.0\%} (63.3 vs.\ 59.3) and AP$_{75}$ by
\textbf{+9.7\%} (77.5 vs.\ 67.8).
The larger AP$_{75}$ gain supports improved mask matching
under the stricter overlap criterion.

\textbf{RWCellIns.}
SRPR-Net improves mAP by \textbf{+3.9\%} (82.4 vs.\ 78.5),
AP$_{50}$ by \textbf{+1.8\%} (96.4 vs.\ 94.6), and
AP$_{75}$ by \textbf{+4.9\%} (94.7 vs.\ 89.8) over
BLO-Inst~\cite{bloinst2026}, extending the performance gains
to microscopic imagery.

\textbf{WheatIns.}
SRPR-Net improves mAP by \textbf{+6.2\%} (74.6 vs.\ 68.4)
and AP$_{75}$ by \textbf{+5.8\%} (89.8 vs.\ 84.0) over
BLO-Inst~\cite{bloinst2026}, despite a 0.2-point lower
AP$_{50}$.
Overall, these results support consistent effectiveness
across the three evaluated visual domains and improved
segmentation at stricter overlap thresholds.

As shown in Table~\ref{tab:sota_penn}, SRPR-Net uses
40.57M trainable parameters versus 38.66M for BLO-Inst,
an increase of only 1.91M. Moreover, SRPR-Net incurs approximately 10\% additional
training time relative to BLO-Inst
(0.057 vs.\ 0.052 GPU hours).

\newcommand{\downbest}[1]{%
  \textbf{#1}\,\rlap{$\downarrow$}%
}

\newcommand{\downsecond}[1]{%
  \underline{#1}\,\rlap{$\downarrow$}%
}

\begin{table}[t]
\centering
\caption{Comparison with state-of-the-art methods on PennFudanPed.
\textbf{Bold} indicates the best results, and \underline{underlined values}
indicate the second-best results.\textsuperscript{1}}
\label{tab:sota_penn}

\footnotesize
\setlength{\tabcolsep}{1.5pt}
\renewcommand{\arraystretch}{1.12}

\begin{tabularx}{\columnwidth}{
@{}
>{\raggedright\arraybackslash}p{0.25\columnwidth}
>{\centering\arraybackslash}p{0.13\columnwidth}
@{\hspace{9pt}}
*{3}{C}
>{\centering\arraybackslash}p{0.13\columnwidth}
>{\centering\arraybackslash}p{0.16\columnwidth}
@{}
}
\toprule

\multirow{3}{*}{Method}
& \multirow{3}{*}{Venue}
& \multicolumn{3}{c}{PennFudanPed}
& \makecell{Trainable}
& \makecell{Train Cost} \\

\cmidrule(lr){3-5}

& &
mAP
& AP$_{50}$
& AP$_{75}$
& Param(M)
& \mbox{(GPU hours)} \\

\midrule

\mbox{Mask R-CNN~\cite{he2017mask}}
& \mbox{ICCV 17}
& 46.4 & 81.8 & 49.0
& 41.48
& \downbest{0.046} \\

\mbox{SOLO~\cite{wang2021solo}}
& \mbox{ECCV 20}
& 42.3 & 75.7 & 39.7
& 46.01
& \underline{0.051} \\

\mbox{SAM+B~\cite{kirillov2023segment}}
& \mbox{ICCV 23}
& 54.3 & 80.5 & 63.7
& 111.08
& 0.150 \\

\mbox{SAM+M~\cite{kirillov2023segment}}
& \mbox{ICCV 23}
& 54.1 & 78.9 & 61.3
& 112.18
& 0.300 \\

\mbox{RS+Anchor~\cite{chen2024rsprompter}}
& \mbox{TGRS 24}
& 41.1 & 68.4 & 45.9
& 117.05
& \underline{0.051} \\

\mbox{RS+Query~\cite{chen2024rsprompter}}
& \mbox{TGRS 24}
& 53.3 & 76.3 & 62.4
& 100.99
& 0.171 \\

\mbox{USIS~\cite{lian2024usis}}
& \mbox{ICML 24}
& 55.1 & 72.6 & 61.0
& 57.02
& 0.081 \\

\mbox{UN-SAM~\cite{chen2025unsam}}
& \mbox{MedIA 25}
& 47.1 & 64.2 & 52.9
& 132.45
& 0.106 \\

\mbox{BLO-Inst~\cite{bloinst2026}}
& \mbox{TMLR 26}
& \underline{59.3}
& \underline{86.9}
& \underline{67.8}
& \downbest{38.66}
& 0.052 \\

\midrule

\textbf{SRPR-Net}
& \textbf{Ours}
& \textbf{63.3}
& \textbf{88.6}
& \textbf{77.5}
& \downsecond{40.57}
& 0.057 \\

\bottomrule
\end{tabularx}
\end{table}
\footnotetext[1]{All training costs are reported in GPU hours and were measured on an NVIDIA GeForce RTX 3090.}

\begin{table}[t]
\centering
\caption{Comparison with state-of-the-art methods on RWCellIns and WheatIns.
\textbf{Bold} indicates the best results, and \underline{underlined values}
indicate the second-best results.}
\label{tab:sota_rw_wheat}

\scriptsize
\setlength{\tabcolsep}{1.5pt}
\renewcommand{\arraystretch}{1.12}

\begin{tabularx}
{\columnwidth}
{@{}l c *{3}{C} @{\hspace{0.6em}} *{3}{C}@{}}
\toprule

\multirow{2}{*}{Method}
& \multirow{2}{*}{Venue}
& \multicolumn{3}{c}{RWCellIns}
& \multicolumn{3}{c}{WheatIns} \\

\cmidrule(lr){3-5}
\cmidrule(lr){6-8}

& & mAP & AP$_{50}$ & AP$_{75}$
& mAP & AP$_{50}$ & AP$_{75}$ \\

\midrule

Mask R-CNN~\cite{he2017mask}
& ICCV 17
& 63.2 & 90.5 & 76.7
& 48.8 & 81.9 & 64.8 \\

SOLO~\cite{wang2021solo}
& ECCV 20
& 64.7 & 89.5 & 74.6
& 58.8 & 91.5 & 71.3 \\

SAM+B~\cite{kirillov2023segment}
& ICCV 23
& 76.1 & 91.6 & 88.2
& 59.2 & 87.9 & 79.6 \\

SAM+M~\cite{kirillov2023segment}
& ICCV 23
& 75.0 & 91.8 & 84.7
& 60.4 & 88.1 & 75.9 \\

RS+Anchor~\cite{chen2024rsprompter}
& TGRS 24
& 76.1 & 92.5 & 88.1
& 62.9 & 87.0 & 78.3 \\

RS+Query~\cite{chen2024rsprompter}
& TGRS 24
& 71.5 & 90.4 & 83.4
& 63.7 & 86.1 & 79.7 \\

USIS~\cite{lian2024usis}
& ICML 24
& 74.3 & 90.8 & 85.2
& 62.5 & 88.5 & 80.3 \\

UN-SAM~\cite{chen2025unsam}
& MedIA 25
& 33.3 & 44.9 & 36.5
& 45.0 & 67.8 & 48.2 \\

BLO-Inst~\cite{bloinst2026}
& TMLR 26
& \underline{78.5}
& \underline{94.6}
& \underline{89.8}
& \underline{68.4}
& \textbf{95.6}
& \underline{84.0} \\

\midrule

\textbf{SRPR-Net}
& \textbf{Ours}
& \textbf{82.4}
& \textbf{96.4}
& \textbf{94.7}
& \textbf{74.6}
& \underline{95.4}
& \textbf{89.8} \\

\bottomrule
\end{tabularx}
\end{table}

\subsection{Ablation Studies}
\label{sec:ablation}

\begin{table}[t]
\centering
\caption{Ablation study on semantic prompt enhancement and
relational prompt refinement on PennFudanPed. The best results
are shown in bold.}
\label{tab:ablation}
\footnotesize
\renewcommand{\arraystretch}{1.12}
\begin{tabular*}{\columnwidth}
{@{\extracolsep{\fill}}lccccc@{}}
\toprule
Method
& Semantic
& Relational
& mAP
& AP$_{50}$
& AP$_{75}$ \\
\midrule
Baseline
& $\times$
& $\times$
& 59.32
& 86.89
& 67.84 \\

+ Semantic
& $\checkmark$
& $\times$
& 62.19
& 85.98
& 75.42 \\

Full SRPR-Net
& $\checkmark$
& $\checkmark$
& \textbf{63.27}
& \textbf{88.58}
& \textbf{77.47} \\
\bottomrule
\end{tabular*}
\end{table}

\begin{figure}[t]
    \centering
    \includegraphics[width=\columnwidth]{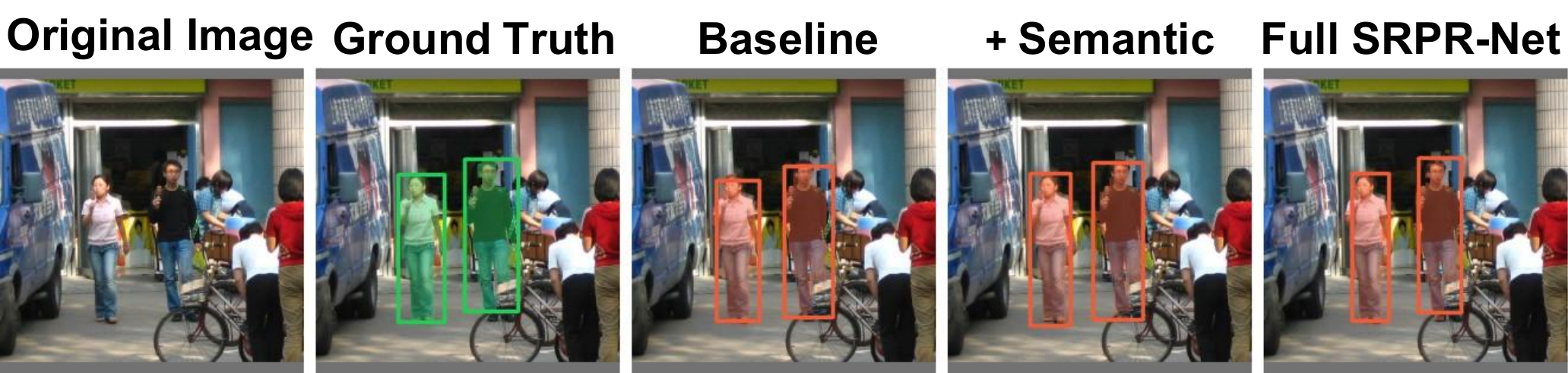}
    \caption{Qualitative ablation on PennFudanPed. From left to right: original image, ground truth, baseline, + Semantic, and full SRPR-Net.}
    \label{fig:ablation_qualitative}
\end{figure}

Table~\ref{tab:ablation} reports an incremental ablation
following the sequential design of SRPR-Net. Because relational refinement operates on the fused tokens produced by semantic enhancement, we evaluate its additional contribution on top of that stage rather than as a standalone component. Fig.~\ref{fig:ablation_qualitative} provides the corresponding qualitative comparisons.

\textbf{Semantic Prompt Enhancement.} Adding CLIP-based semantic enhancement yields absolute gains of +2.87\% in mAP and \textbf{+7.58\%} in AP$_{75}$, despite a 0.91\% decrease in AP$_{50}$. These results suggest improved mask alignment at the stricter IoU threshold.

\textbf{Relational Prompt Refinement.} Adding relational refinement to the semantic tokens yields further absolute gains of +1.08\%, +2.60\%, and +2.05\% in mAP, AP$_{50}$, and AP$_{75}$, respectively, supporting the benefit of inter-instance modeling beyond semantic enhancement.
The smaller incremental gains may reflect the information
already provided by the semantic tokens, which combine
geometric embeddings with CLIP visual and text features.
Moreover, the shared global visual feature and common
pedestrian text feature provide limited additional semantic
diversity for relational modeling.
Since both stages refine existing detector proposals, missed instances cannot be recovered by the current framework. Overall, the ablation indicates that semantic enhancement provides the primary gain, while relational refinement offers a smaller but consistent improvement.

\section{Conclusion}
\label{sec:conclusion}
In this work, we present SRPR-Net, a novel framework for automated SAM-based instance segmentation. By integrating CLIP-based semantic enhancement with Transformer-based relational refinement, our method enriches detector-generated prompts with visual-language semantics and inter-instance context. Experimental results demonstrate competitive performance against state-of-the-art methods and support the effectiveness of the refinement strategy. In the future, we will investigate prompt refinement under missed or inaccurate detections and evaluate transfer to unseen domains.

\clearpage          
\bibliographystyle{IEEEbib}
\bibliography{refs}

\end{document}